\def\paperlanguage{}
\pdfoutput=1 % for arxiv

\documentclass[letterpaper, 10 pt, conference]{ieeeconf}  % Comment this line out if you need a4paper

\usepackage{bm}
\usepackage{cite}
\usepackage{flushend}
\include{preamble}

\IEEEoverridecommandlockouts                              % This command is only needed if
\title{\LARGE \textbf
  {
    \switchlanguage%
    {%
      Realization of Seated Walk by a Musculoskeletal Humanoid with Buttock-Contact Sensors From Human Constrained Teaching
    }%
    {%
      人間の制約付き教示による臀部接触センサを有する筋骨格ヒューマノイドの座り歩きの実現
    }%
  }
}

\author{Kento Kawaharazuka$^{1}$, Kei Okada$^{1}$, and Masayuki Inaba$^{1}$% <-this % stops a space
  \thanks{$^{1}$ The authors are with the Department of Mechano-Informatics, Graduate School of Information Science and Technology, The University of Tokyo, 7-3-1 Hongo, Bunkyo-ku, Tokyo, 113-8656, Japan.
    {\texttt\small [kawaharazuka, k-okada, inaba]@jsk.t.u-tokyo.ac.jp}
  }
}

\begin{document}

\maketitle
\thispagestyle{empty}
\pagestyle{empty}

%%%%%%%%%%%%%%%%%%%%%%%%%%%%%%%%%%%%%%%%%%%%%%%%%%%%%%%%%%%%%%%%%%%%%%%%%%%%%%%%
\begin{abstract}
  \switchlanguage%
  {%
    In this study, seated walk, a movement of walking while sitting on a chair with casters, is realized on a musculoskeletal humanoid from human teaching.
    The body is balanced by using buttock-contact sensors implemented on the planar interskeletal structure of the human mimetic musculoskeletal robot.
    Also, we develop a constrained teaching method in which one-dimensional control command, its transition, and a transition condition are described for each state in advance, and a threshold value for each transition condition such as joint angles and foot contact sensor values is determined based on human teaching.
    Complex behaviors can be easily generated from simple inputs.
    In the musculoskeletal humanoid MusashiOLegs, forward, backward, and rotational movements of seated walk are realized.
  }%
  {%
    本研究では, 人間がキャスター付きの椅子に座ったまま移動する動作であるseated walkを, 筋骨格ヒューマノイドにおいて, 人間の教示から実現する.
    人体模倣構造における面状骨格間構造に実装した尻の接触センサを用いて身体のバランスを取る.
    また, 一次元制御入力とその遷移, 遷移条件を予め記述し, 人間による教示の際の状況から, 遷移条件である関節角度や足の接触センサ値の閾値を決定するような制約付き教示手法を開発する.
    シンプルな入力から複雑な動作を簡単に生成することができる.
    筋骨格ヒューマノイドMusashiOLegsにおいて, 座り歩きにより前進後退回旋運動を実現した.
  }%
\end{abstract}

\section{Introduction}\label{sec:introduction}
\begin{figure}[t]
  \centering
  \includegraphics[width=0.95\columnwidth]{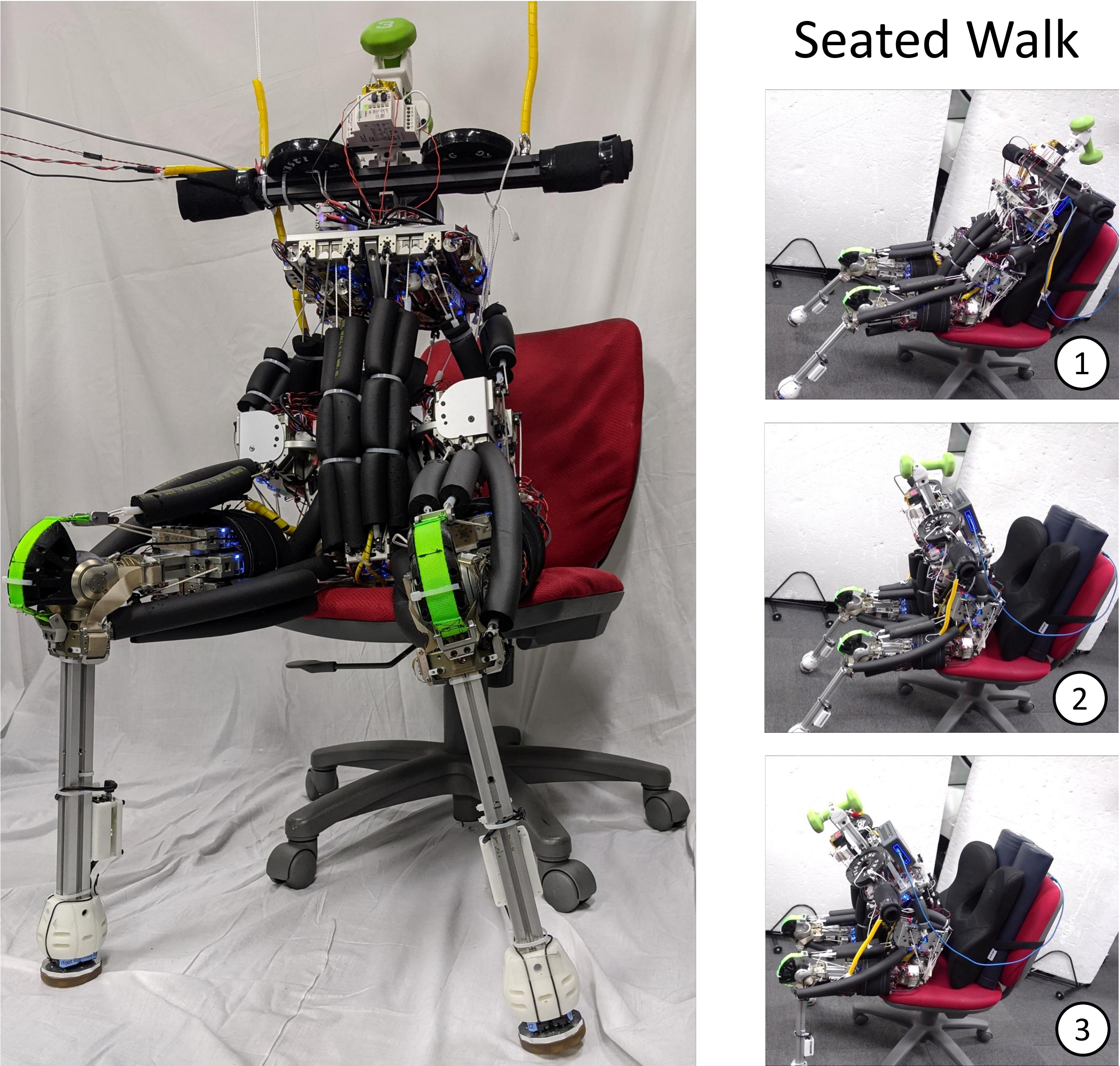}
  \caption{Seated walk by the musculoskeletal humanoid MusashiOLegs.}
  \label{figure:concept}
\end{figure}

\switchlanguage%
{%
  Until now, the field of bipedal humanoids has focused mainly on standing and walking in terms of locomotion \cite{kajita2001walking, hirose2007asimo}.
  Without the help of the environment, robots walk generally based on theories such as zero moment point \cite{kajita2003zmp} and capture point \cite{pratt2006capture}.
  For robots, locomotion using the environment such as walking upstairs using a handrail \cite{harada2004handrail, werner2015multicontact} is more difficult than the usual walking.
  On the other hand, for humans, it is easier to move in closer contact with the environment, such as walking while grasping a handrail or moving while crawling \cite{adolph1997infant}.
  This is due to the errors in recognizing and modeling the environment, the lack of sensors to measure contact with the environment, and the fact that the environment becomes a constraint due to the rigidity of the robot.
  In this study, we focus on a movement called seated walk, which is an example of a movement with environmental contact, in which a person sits on a chair with casters and moves to pick up documents, phone, etc. (\figref{figure:concept}).

  Among humanoids, the musculoskeletal humanoid \cite{gravato2010ecce1, nakanishi2013design, kawaharazuka2019musashi} has a body shape and actuation mechanism that are closer to those of a human, and it has flexibility in its body due to the elongation of muscle wires, nonlinear elastic elements, and soft foam covers.
  Among these, the newly developed musculoskeletal humanoid MusashiOLegs \cite{onitsuka2021musashiolegs} has planar interskeletal structures all over the body, which enables the realization of environmental contact on a wider surface and stable muscle routes not by using each thin muscle wire but by making the wires planar.
  Using the planar interskeletal structure of the buttocks of MusashiOLegs, we implement buttock-contact sensors.
  The purpose of this research is to realize seated walk with MusashiOLegs, a musculoskeletal humanoid that has the same proportions and flexibility as a human and can measure the contact between the body and environment.
}%
{%
  これまで二足歩行型のヒューマノイドの分野では, 移動については主に立って歩くというところに焦点が当てられてきた\cite{kajita2001walking, hirose2007asimo}.
  環境の助けを借りずに, zmp \cite{kajita2003zmp}やcapture point \cite{pratt2006capture}等の理論を用いて歩行させる技術が一般的である.
  手すりを使った階段歩行のような, 環境を使って移動する技術\cite{harada2004handrail, werner2015multicontact}の方がロボットにとってはより難しいというのが現状である.
  これに対して, 人間にとっては, 手すりを掴みながら歩く, 椅子に座ったまま歩いて移動する, ハイハイしながら移動するといった, より環境と密に接触するような移動動作の方が簡単である\cite{adolph1997infant}.
  これは, 環境の認識・モデル化誤差や, 環境との接触を測るセンサがないこと, ロボットの固さゆえに環境が拘束となってしまうこと等に由来する.
  本研究では, この環境接触を伴う移動の一例である, キャスター付きの椅子に座ったまま, 書類や電話をを取ったり, 話したり等のため移動する, seated walkと呼ばれる動作に着目する(\figref{figure:concept}).

  ヒューマノイドの中でも, 筋骨格ヒューマノイド\cite{gravato2010ecce1, nakanishi2013design, kawaharazuka2019musashi}は人間により近い身体形状・駆動方法を有しており, 筋ワイヤや非線形弾性要素の伸びにより身体に柔軟性が備わっている.
  この中でも, 新しく開発したMusashiOLegs \cite{onitsuka2021musashiolegs}は身体の至るところに面状骨格間構造を有しており, 筋ワイヤ一本ずつで身体を駆動・拘束するのではなく, それらを面状にすることで, より広い面での環境接触・安定的な筋経路の実現が可能となっている.
  そして, このMusashiOLegsのお尻の面状骨格間構造を利用し, お尻接触センサを実装する.
  本研究の目的は, この人間と同じプロポーション・身体柔軟性を持ち, 環境である椅子との間の接触を知ることが可能な, 筋骨格ヒューマノイドMusashiOLegsによりseated walkを実現することである.
}%

\switchlanguage%
{%
  In previous researches, motions such as sitting on a chair \cite{noda2015contact}, standing up from a chair, and moving on a desk \cite{fukazawa2020multicontact} have been performed.
  In these cases, the contact force is estimated by using six-axis force sensors and IMUs in the hands and feet, and the motion is generated from the kinematic and dynamic models.
  However, musculoskeletal humanoids, which have a flexible structure similar to that of humans and are difficult to modelize, need to acquire motions through human teaching and learning.
  Although there are several teaching methods using a bilaterally controllable device \cite{ishiguro202tablis}, VR device \cite{zhang2018imitation} or motion capture \cite{stanton2013teleoperation}, these methods require additional devices and, as in the task of seated walk, these methods are not suitable for cases where careful teaching is desirable because the robot cannot be recovered once its balance is lost.
  Also, there are studies that directly measure the contact force by attaching contact sensors around the whole body and control the movement \cite{ohmura2007tactile, kumagai2012tactile}.
  On the other hand, in musculoskeletal humanoids, it is difficult to mount a contact sensor on the body surface because the muscles are arranged to wrap around the body.
  \cite{osada2012planar} measures the lateral force applied to the muscles as muscle tension, but it cannot separate the contact force from the movement of muscles and thus cannot correctly measure the contact between the chair and buttocks.

  From these points of view, in this study, we develop a buttock-contact sensor where the planar interskeletal structure of the buttocks of musculoskeletal humanoids is utilized and contact sensors are inserted into it.
  Using this sensor, we implement a basic balance controller of the body.
  Also, we develop a constrained teaching method (CTM) in which one-dimensional control command, its transition, and transition condition at each state are described in advance, and only the threshold value of each transition condition such as joint angles and foot contact sensor values are learned from human teaching.
  By limiting the control command for teaching to one dimension, we can simplify the teaching process and make it possible to teach even complex motions only with a slide bar on the screen.
  In addition, from the transitions of the control commands during the teaching, we determine the threshold values for the transition conditions and reproduce the motions using these values.
  Since only the threshold value is learned, the system can only handle quasi-static motions.
  On the other hand, even in the case of slow and careful teaching, the reproduction speed of the taught motion can be set arbitrarily for each state.
  Combining these methods, we have realized a seated walk using the musculoskeletal humanoid MusashiOLegs.

  The contribution of this research is as follows.
  \begin{itemize}
    \item Development of contact sensors on the planar interskeletal structure of the buttocks and a balance control using them
    \item Implementation of a constrained teaching method that generates complex motions from simple inputs by learning only the threshold of transition conditions
    \item Realization of seated walk by a musculoskeletal humanoid
  \end{itemize}

  The structure of this research is as follows.
  In \secref{sec:hardware}, we introduce the musculoskeletal humanoid MusashiOLegs, its planar interskeletal structure, and the implementation of the buttock-contact sensor.
  In \secref{sec:proposed-method}, we describe the constrained teaching method, the balance control using the buttock-contact sensor, and the whole system.
  In \secref{sec:experiments}, we experiment with forward, backward, and rotational movements, and balance control in seated walking, and show its effectiveness through an integration experiment.
  In \secref{sec:discussion}, we discuss the experiments and the limitation of this study, and conclude in \secref{sec:conclusion}.
}%
{%
  これまでの先行研究では, 椅子に座る, 椅子から立ち上がるような動作\cite{noda2015contact}, 腰掛けた机の上を移動するような動作\cite{fukazawa2020multicontact}等が行われている.
  これらは手や足の六軸センサ・IMU等を用いて接触力を推定し, 運動モデルから動作を生成する.
  しかし, より人間に近く柔軟な構造を持つモデル化の難しい筋骨格ヒューマノイドにおいてこれらは難しく, 人間の教示や学習によって動作を獲得していかなければならない.
  教示手法としては, バイラテラル制御可能なデバイス\cite{ishiguro202tablis}, VR デバイス \cite{zhang2018imitation}やmotion capture \cite{stanton2013teleoperation}を利用した方法等があるが, これらは追加のデバイスが必要かつ, 本タスクのように一度バランスを崩すと元に戻ることが難しく慎重に動作を教示したいケースには適していない.
  また, 全身に接触センサをまとわせ, 接触力を直接測定し制御する研究も存在する\cite{ohmura2007tactile, kumagai2012tactile}.
  一方で, 筋骨格ヒューマノイドは現状骨格に巻き付くように筋が配置されるため, 身体表面上に接触センサを搭載することが難しい.
  \cite{osada2012planar}は筋に対して横からかかる力を筋張力として測定するが, 自身の動きと接触力を分離することができず, 椅子と尻の接触を正しく計測できない.

  これらの観点から本研究では, 筋骨格ヒューマノイドの臀部の面状骨格間構造を利用し, この中に接触センサを実装した, 臀部接触センサを開発する.
  これを使い, 身体の基本的なバランス制御を実装する.
  また, 一次元制御入力とその遷移, 遷移条件を予め記述し, 人間による教示の際の状況から, 遷移条件である関節角度や足の接触センサ値の閾値のみを学習するような, 制約付きの教示方法を開発する.
  教示の際の制御入力を一次元に絞ることで教示をよりシンプルにし, 複雑な動作でも画面上のスライドバーのみで教示を行うことができる.
  また, 教示の際の制御入力の遷移から, 遷移条件における閾値を決定し, これを使って動作を再生する.
  閾値のみについて学習を行うため, 準静的な動作しか扱うことはできないが, ゆっくり慎重に教示した場合も, 動作速度については別に設定するため, 教示動作の再生速度は遷移ごとに任意に設定することができる.
  これらの手法を組み合わせ, 筋骨格ヒューマノイドMusashiOLegsによるseated walkを実現した.

  本研究の貢献は以下となっている.
  \begin{itemize}
    \item お尻の面状骨格間構造への接触センサの実装とバランス制御
    \item 遷移条件の閾値のみを学習することで複雑な動作をシンプルな入力から生成可能な制約付き教示方法の開発
    \item 筋骨格ヒューマノイドにおけるseated walkの実現
  \end{itemize}

  本研究の構成は以下のようになっている.
  \secref{sec:hardware}では, MusashiOLegsの紹介とその面状骨格間構造, お尻接触センサの実装について述べる.
  \secref{sec:proposed-method}では, 制約付き教示手法とお尻センサによるバランス制御, 全体システムについて述べる.
  \secref{sec:experiments}では, seated walkにおける前進・後退・回旋・バランス制御について実験し, 統合実験によりその有効性を示す.
  \secref{sec:discussion}では, 実験に関する議論・本研究のlimitationを述べ, \secref{sec:conclusion}で結論を述べる.
}%

\begin{figure}[t]
  \centering
  \includegraphics[width=0.95\columnwidth]{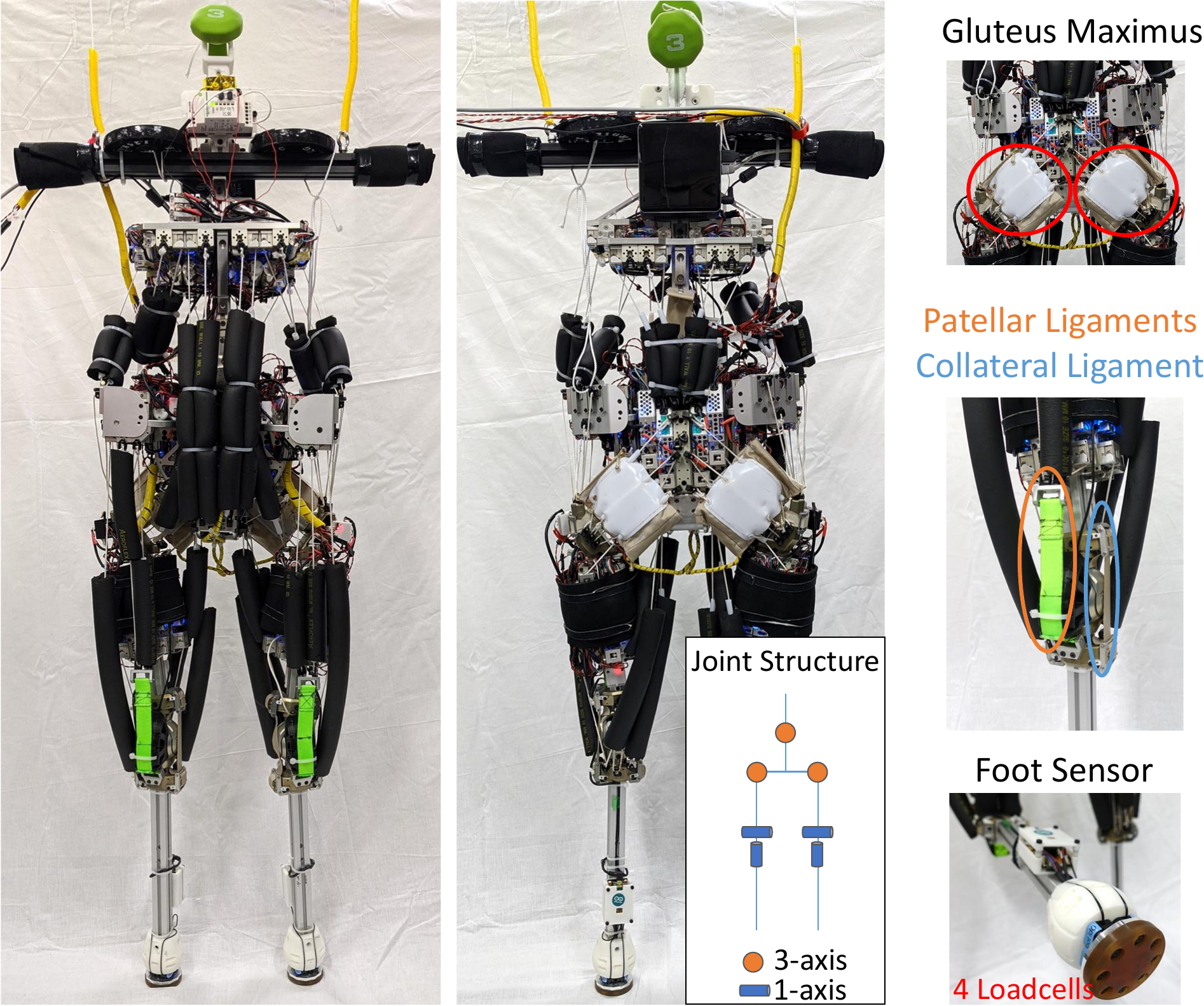}
  \caption{The musculoskeletal humanoid MusashiOLegs \cite{onitsuka2021musashiolegs} with various planar interskeletal structures.}
  \label{figure:musashiolegs}
\end{figure}

\section{The Musculoskeletal Humanoid MusashiOLegs and Implementation of Buttock-Contact Sensors}\label{sec:hardware}
\subsection{Overview of MusashiOLegs} \label{subsec:musashiolegs}
\switchlanguage%
{%
  The overall view of the musculoskeletal humanoid MusashiOLegs \cite{onitsuka2021musashiolegs} used in this study is shown in the left figures of \figref{figure:musashiolegs}.
  Usually, in musculoskeletal humanoids, muscles are redundantly arranged around the joints.
  MusashiOLegs uses an electric motor and a pulley to wind the muscle wires, though there are various methods to drive the muscles.
  Muscle temperature $c$, muscle tension $f$, and muscle length $l$ can be measured from the muscle module, which consists of a motor, pulley, circuit, and sensors.
  In the musculoskeletal structure, although joint angles cannot usually be measured directly due to the presence of the complex spine and spherical joints, etc., joint angles of MusashiOLegs can be directly measured using pseudo ball joints \cite{kawaharazuka2019musashi}.
  % Even if direct measurement is not possible, it is possible to estimate the joint angle from muscle length and muscle tension \cite{kawaharazuka2019longtime}.
  In MusashiOLegs, only 13 joints are implemented.
  Each joint is denoted as T-r, T-p, T-y, lH-r, lH-p, lH-y, lK-p, lK-y, rH-r, rH-p, rH-y, rK-p, and rK-y (where T is Torso, H is Hip, K is Knee, $\{l, r\}$ is left and right, and $\{r, p, y\}$ is roll, pitch, and yaw).
  The joint angle of $joint$ is represented by $\theta_{joint}$.
  Four loadcells are distributed at the tip of each foot to measure the contact force.
  In this study, $F_{\{lfoot, rfoot\}}$ is the sum of the values of the loadcells of each foot, and $F_{foot}$ is $F_{lfoot}+F_{rfoot}$.
  Since the muscle wires are made of Dyneema\textsuperscript{\textregistered}, an abrasion resistant synthetic fiber, and are surrounded by a soft foam cover, their elasticity provides the flexibility of the body.
  By learning the relationship between muscle length, muscle tension, and joint angle, it is possible to control the joint angle \cite{kawaharazuka2018online, kawaharazuka2019longtime, kawaharazuka2020autoencoder}.
  However, due to the effects of friction and hysteresis, it is not always possible to control the joint angle accurately enough.
  Therefore, the measured or estimated joint angle and the commanded joint angle are often different, and the expression $\theta_{joint}$ in this study refers to the commanded joint angle.

  We describe the planar interskeletal structure, which is a unique feature of MusashiOLegs.
  Normally, muscles are driven by thin wires in musculoskeletal humanoids, but in MusashiOLegs, the planar interskeletal structure is adopted at several points.
  As shown in the right figures of \figref{figure:musashiolegs}, the collateral ligament and patellar ligament of the knee are constructed in a planar structure, which realizes a screw home mechanism and a large moment arm.
  In addition, the gluteus maximus muscle is implemented by a planar structure that runs through multiple wires, ensuring a large moment arm and flexibility in environmental contact.
}%
{%
  本研究で用いる筋骨格ヒューマノイドMusashiOLegs \cite{onitsuka2021musashiolegs}の全体図を\figref{figure:musashiolegs}の左図に示す.
  通常, 筋骨格ヒューマノイドにおいては, 関節の周りに冗長に筋が配置されている.
  筋の駆動方法には様々な形式があるが, MusashiOLegsは電気モータとプーリにより筋ワイヤを巻き取る方式を採用している.
  モータやプーリ・回路・センサ等が一体となった筋モジュールからは, 筋温度$c$, 筋張力$f$,　筋長$l$を測定することが可能である.
  筋骨格構造においては, 複雑な背骨や球関節等が存在するため, 通常関節角度を直接測定することはできないが, MusashiOLegsの関節は擬似球関節を有しており, 関節角度を直接測定することができる\cite{kawaharazuka2019musashi}.
  % 直接測定することができない場合でも, 筋長と筋張力から, 関節角度を推定することが可能である\cite{kawaharazuka2019longtime}.
  MusashiOLegsにおいて関節は背骨・腰・膝の13関節のみ実装されている.
  それぞれの関節はT-r, T-p, T-y, lH-r, lH-p, lH-y, lK-p, lK-y, rH-r, rH-p, rH-y, rK-p, rK-yと表す(ここで, TはTorso, HはHip, KはKnee, $\{l, r\}$はleftとright, $\{r, p, y\}$はroll, pitch, yawを表す).
  $\theta_{joint}$により$joint$の関節角度を表現する.
  足の先端にロードセルが4つずつ分布しており, 接触力を測定することが可能である.
  本研究では, $F_{\{lfoot, rfoot\}}$を左足・右足のそれぞれのロードセルの値の合計値, $F_{foot}$を$F_{lfoot}+F_{rfoot}$とする.
  筋ワイヤには摩擦に強い合成繊維であるDyneema\textsuperscript{\textregistered}が使用されており, その周りに発泡性カバーが取り付けられているため, それらの弾性により柔軟性が生まれる.
  筋長・筋張力・関節角度の間の関係を学習することで, 関節角度を制御することが可能であるが\cite{kawaharazuka2018online, kawaharazuka2019longtime, kawaharazuka2020autoencoder}, 摩擦・ヒステリシスの影響から, 十分正確に制御できるわけではない.
  そのため, 実測または推定される関節角度と指令関節角度は異なる場合が多く, 本研究において$\theta_{joint}$という表現は指令関節角度を指すこととする.

  MusashiOLegsに特有の特徴である面状骨格間構造について述べる.
  通常筋骨格ヒューマノイドにおいて筋は細いワイヤによって駆動されるが, MusashiOLegsにおいては, 複数の箇所で面状の構造を採用している.
  \figref{figure:musashiolegs}の右図に示すように, 膝の側副靭帯と膝蓋骨靭帯が面状に構築されており, 終末強制回旋機構と大きなモーメントアームを実現している.
  また, 大殿筋は複数のワイヤを通るような面状構造によって実装されており, 高いモーメントアーム維持力と環境接触の柔軟性を担保する.
}%

\begin{figure}[t]
  \centering
  \includegraphics[width=0.95\columnwidth]{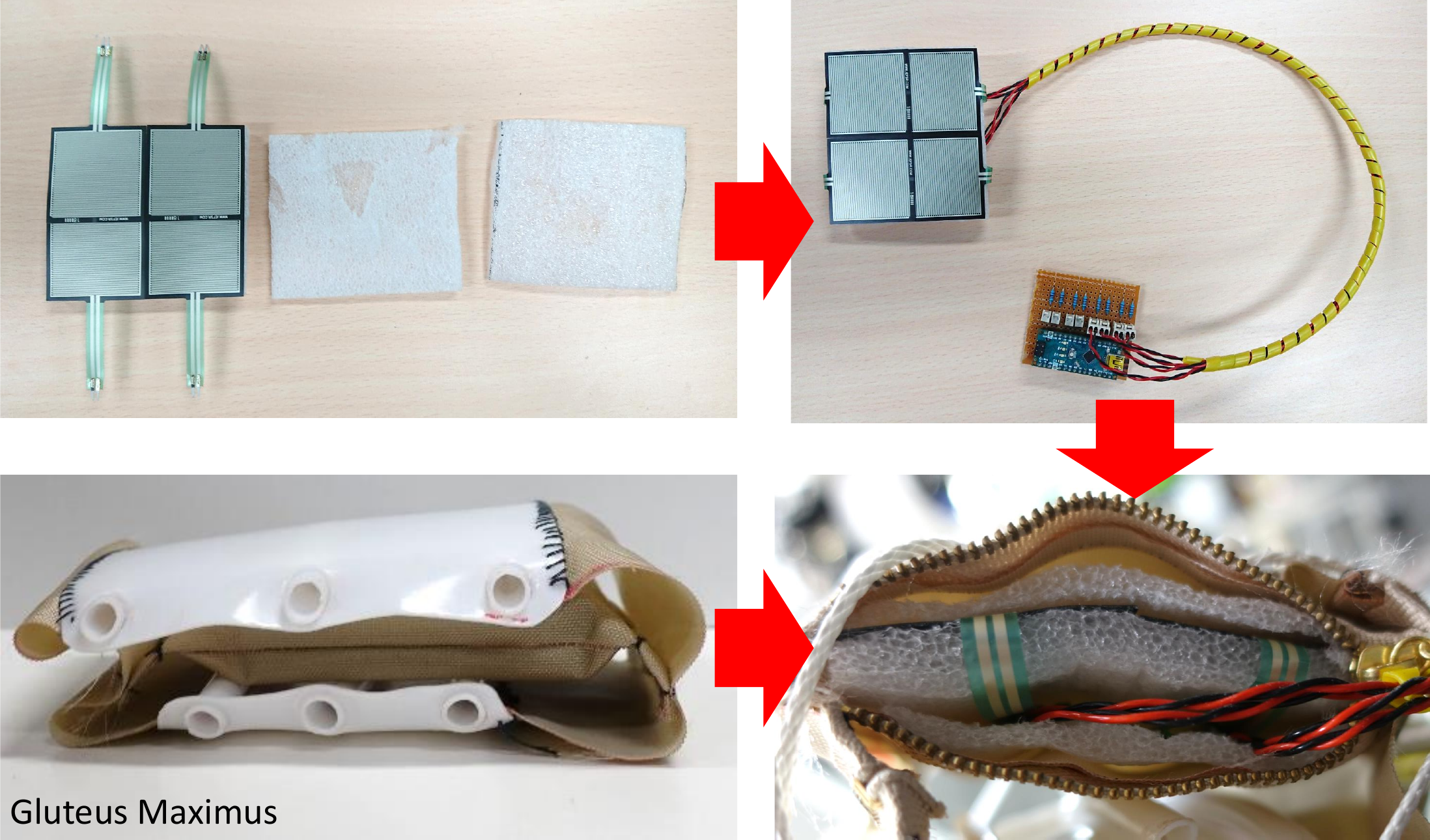}
  \caption{The implementation of buttock-contact sensors in the planar interskeletal structure of gluteus maximus.}
  \label{figure:hip-sensor}
\end{figure}

\begin{figure}[t]
  \centering
  \includegraphics[width=0.95\columnwidth]{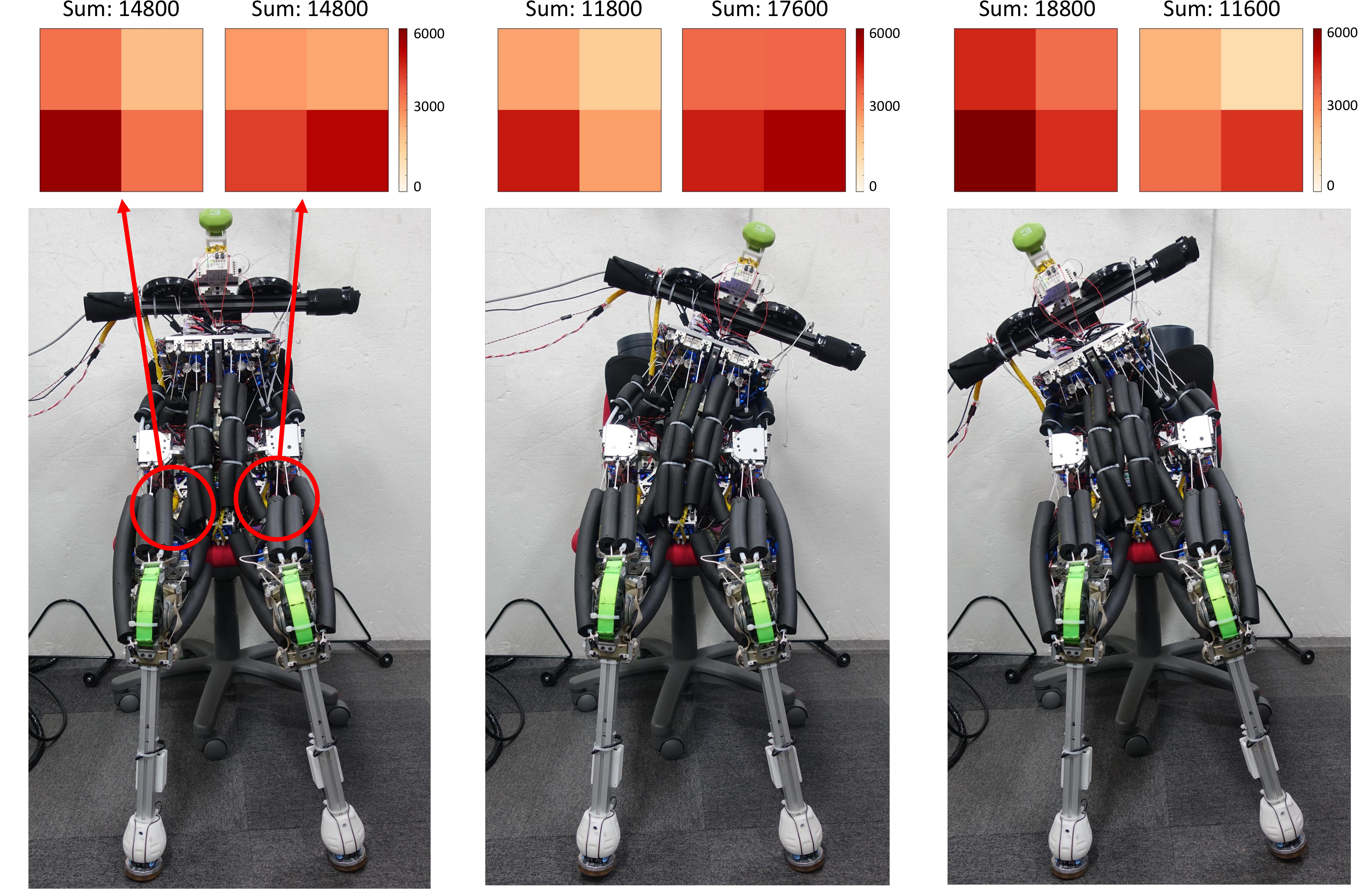}
  \caption{The difference in buttock-contact forces with various ways of sitting.}
  \label{figure:hip-sensor-exp}
\end{figure}

\subsection{Buttock-Contact Sensor} \label{subsec:hip-sensor}
\switchlanguage%
{%
  The implementation of the buttock-contact sensor is shown in \figref{figure:hip-sensor}.
  Three thin sheets of foam material are prepared, and four large pressure sensors FSR\textsuperscript{\textregistered}406 are attached to one of them.
  It is sandwiched between the other two sheets and inserted into the planar interskeletal structure of the gluteus maximus muscle.
  The cable passes through the pubic symphysis and the analog value is measured by Arudino near the sacrum.
  Since the sensitivity of FSR is nonlinear and the larger the force, the smaller the potential difference becomes, we use the corrected value as $\exp(A/100)$.
  Here, $A$ is a 10-bit analog value, and the pressure sensor has a wide contact surface, so the exact magnitude of the force cannot be measured.
  $F_{\{lhip, rhip\}}$ is the sum of the four contact sensor values for each left or right buttock.
  As shown in \figref{figure:hip-sensor-exp}, we can see that there is a difference in the values of the buttock-contact sensors depending on how the robot sits on the chair.
  In parallel with CTM generating the overall motion, the body balance is controlled by using the buttock-contact sensors.
}%
{%
  本研究で実装した, お尻接触センサの実装を\figref{figure:hip-sensor}に示す.
  薄い発泡性素材を3枚用意し, その一つに4つの大きな圧力センサFSR\textsuperscript{\textregistered}406を装着する.
  残りの2枚でそれをサンドイッチし, 大殿筋の面状骨格間構造内に挿入する.
  ケーブルは恥骨結合部を通り, 仙骨付近のArudinoによってアナログ値を計測する.
  FSRの感度は非線形であり, 力が大きいほど電位差は小さくなってしまうため, 本研究では$exp(A/100)$のような形で補正した値を用いる.
  なお, $A$は10 bitのアナログ値であり, 圧力センサは広い接触面を持つため, 正確な力の大きさはわからない.
  $F_{\{lhip, rhip\}}$は左尻, 右尻のそれぞれの4つの接触センサ値の合計値とする.
  \figref{figure:hip-sensor-exp}のように, 椅子への座り方によってお尻接触センサの値に違いが出ていることがわかる.
  CTMが全体の動作を生成するのと並行に, このお尻接触センサにより体のバランス制御を行う.
}%

\begin{figure*}[t]
  \centering
  \includegraphics[width=1.95\columnwidth]{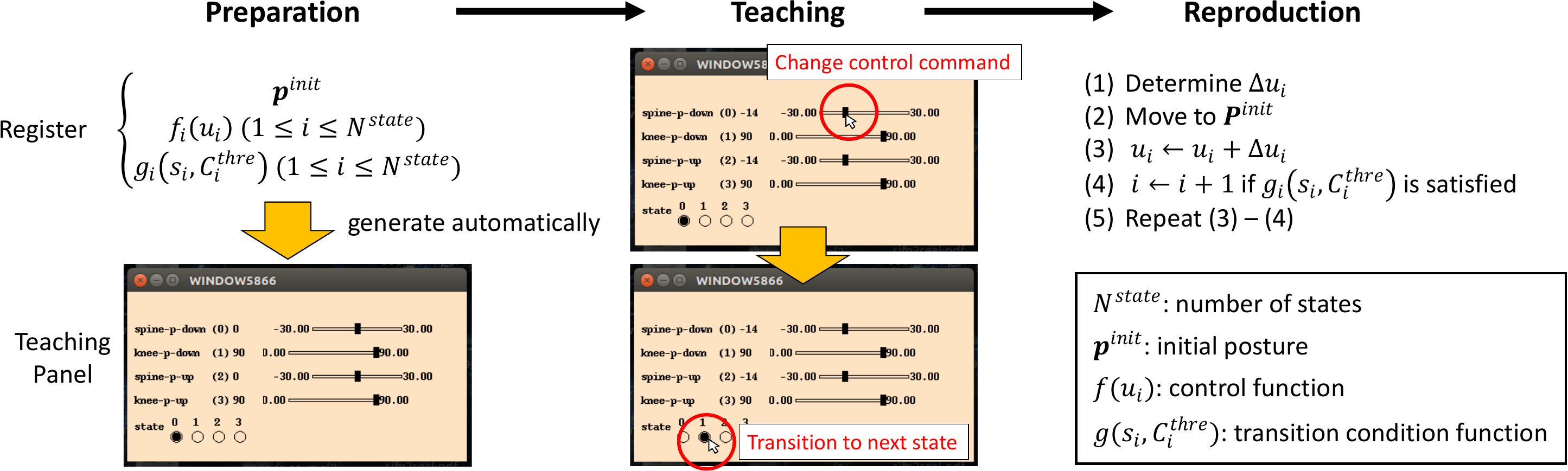}
  \caption{The procedures of constrained teaching method: preparation, teaching, and reproduction.}
  \label{figure:teaching-method}
\end{figure*}

\section{Seated Walk with Constrained Teaching Method and Buttock-Contact Balancer} \label{sec:proposed-method}
\subsection{Constrained Teaching Method} \label{subsec:teaching-method}
\switchlanguage%
{%
  We describe a constrained teaching method (CTM).
  This method differs from the usual teaching method in that the following assumptions are made on the behavior.
  Here, $i$ is the current transition state, $u$ is the one-dimensional control command, $s$ is some sensor state for transition condition, $C^{thre}$ is the threshold of the sensor state, and $\{u, s, C^{thre}\}_{i}$ denotes $\{u, s, C^{thre}\}$ at $i$.
  \begin{itemize}
    \item There is an explicit state transition in the behavior and each control command $u_{i}$ can only be given in one dimension.
    \item The transition condition is expressed by the relationship between $s_{i}$ and $C^{thre}_{i}$, and $s_{i}$ must change depending on $u_{i}$.
    \item the behavior can be reproduced by learning only $C^{thre}$.
  \end{itemize}
  With these assumptions, \figref{figure:teaching-method} shows how preparation, teaching, and reproduction are executed.

  First, the preparation before teaching is as follows.
  \begin{itemize}
    \item Register the initial posture $\bm{p}^{init}$.
    \item Register the control function $f_{i}(u_{i})$ and the transition condition function $g_{i}(s_{i}, C^{thre}_{i})$ ($1\leq i\leq N^{state}$).
  \end{itemize}
  Here, $N^{state}$ is the number of states.
  For example, $f_{i}$ may move the pitch axis of the torso or specify the x-coordinate position in three-dimensional space for inverse kinematics.
  Let $g_{i}$ be a condition such as $s_{i} \geq C^{thre}_{i}$ or $s_{i} \leq C^{thre}_{i}$, which returns True when satisfied.
  This preparation are made by humans while considering the target task.

  Next, the teaching procedure is as follows
  \begin{enumerate}
    \item Transition to the initial posture $\bm{p}^{init}$.
    \item $u_{i}$ is manipulated from the teaching panel.
    \item Stop changing $u_{i}$ at the desired position and transition to $i+1$.
  \end{enumerate}
  Here, the learning process of determining the parameter $C^{thre}_{i}$ takes place.
  The value of $s_{i}$ is registered as $C^{thre}_{i}$ when the state is moved from $i$ to $i+1$.

  Finally, the reproduction procedure is as follows.
  \begin{enumerate}
    \item Determine the change in control command $\Delta{u}_{i}$.
    \item Transition to the initial posture $\bm{p}^{init}$.
    \item Repeat $u_{i}\gets u_{i} + \Delta{u}_{i}$ to change $u_{i}$.
    \item Move to $i+1$ when the transition condition is satisfied.
    \item Repeat 3) -- 4).
  \end{enumerate}

  Although this is a very simple teaching method, it is powerful for certain behaviors where the motion can be quasi-statically described only by transitions of one-dimensional control commands.
  The only term to be learned is the threshold value of $C^{thre}_{i}$, and $\Delta{u}_{i}$ is arbitrarily set for each state before reproduction.
  Therefore, it is possible to operate the robot slowly and carefully so as not to lose its balance, and then reproduce it at a fast speed.
}%
{%
  制約付き教示方法(CTM)について述べる.
  本手法は通常の教示方法とは違い, 動作に以下の仮定を置いたうえで教示を行う.
  ここで, $i$は現在の遷移状態, $u$は一次元制御入力, $s$は何らかのセンサ状態, $C^{thre}$はセンサ状態の閾値, $\{u, s, C^{thre}\}_{i}$は$i$における$\{u, s, C^{thre}\}$を表す.
  \begin{itemize}
    \item 動作に明示的な状態遷移があり, 一つの状態$i$について制御入力$u_{i}$は一次元しか与えることができない.
    \item 状態遷移の遷移条件は$s_{i}$と$C^{thre}_{i}$の大小によって表現され, $u_{i}$によって$s_{i}$が変化しなければならない.
    \item $C^{thre}$のみ学習することで, 動作が再生可能である.
  \end{itemize}
  これらの仮定を置いたうえで, 準備・教示・学習・再生がどのように行われるのかを\figref{figure:teaching-method}に示す.

  まず, 教示前の準備は以下である.
  \begin{itemize}
    \item 初期姿勢$\bm{p}^{init}$の登録
    \item それぞれの遷移における制御関数$f_{i}(u_{i})$, 遷移条件関数$g_{i}(s_{i}, C^{thre}_{i})$の登録 ($1 \leq i \leq N^{state}$)
  \end{itemize}
  ここで, $N^{state}$は状態数とする.
  例えば, $f_{i}$は腰のピッチ軸を動かすものであったり, 三次元空間におけるx座標の位置を指定するものであったりする.
  $g_{i}$は$s_{i} \geq C^{thre}_{i}$や$s_{i} \leq C^{thre}_{i}$のような条件であり, 満たされた時にTrueを返すものとする.
  これらの設定は人間がそのタスクを見ながら設定する.

  次に, 教示の手順は以下である.
  \begin{enumerate}
    \item 初期姿勢$\bm{p}^{init}$へ移行
    \item $u_{i}$を教示ウィンドウから操作
    \item 所望の位置で$u_{i}$の変化を止め, 遷移を$i+1$へ移行
  \end{enumerate}
  ここで, パラメータ$C^{thre}_{i}$の決定という学習が行われる.
  遷移を$i$から$i+1$に移行する際の, $s_{i}$の値を$C^{thre}_{i}$として登録していく.

  最後に, 再生の手順は以下である.
  \begin{enumerate}
    \item 1ステップにおける制御入力の変化$\Delta{u}_{i}$を決定する.
    \item 初期姿勢$\bm{p}^{init}$へ移行
    \item $u_{i} \gets u_{i} + \Delta{u}_{i}$により$u_{i}$を変化させる.
    \item 遷移条件が満たされたら$i+1$へ移行
    \item (3) -- (4)を繰り返す
  \end{enumerate}

  非常にシンプルな教示手法であるが, 一次元制御入力の遷移だけで準静的に動きが記述可能な, 特定の動作においては非常に強力な効果を持つ.
  学習される項は$C^{thre}_{i}$の閾値だけであり, $\Delta{u}_{i}$を再生の前に遷移ごとに自身で設定するため, 教示の際の動作速度に, 再生の際の動作速度は依存しない.
  そのため, バランス等を崩さないように慎重にゆっくりロボットを操作し, 再生の際は速い速度で再生するということが可能になる.
}%

\subsection{Buttock-Contact Balancer} \label{subsec:hip-balancer}
\switchlanguage%
{%
  We use buttock-contact sensors for balancing.
  Because the back of the chair is behind the robot, it is difficult to lose the front-back balance.
  On the other hand, the problem in seated walk is the misalignment of the left and right balance.
  In order to solve this problem, the following PI control is applied to control the tactile balance of the buttocks.
  \begin{align}
    d &= F_{lhip}-F_{rhip}\\
    D &\gets D + d\\
    \theta_{T-r} &= C_{pgain}d + C_{igain}D
  \end{align}
  Here, $C_{\{pgain, igain\}}$ is the gain of the PI control.
  In this study, $C_{pgain}=5.0$ and $C_{igain}$ is set to $C_{igain}=0.3$ for forward and backward motions and $C_{igain}=0.03$ for rotational motions.
}%
{%
  お尻接触センサを用いてバランスを取ることを行う.
  身体の後ろには椅子の背板があるため, 前後のバランスは崩しにくい.
  それに対して, seated walkの際に問題となるのが, 左右のバランスのズレである.
  徐々に片方のお尻に力がのっていき, 横に倒れるという問題を解消するため, 以下のPI制御によるお尻触覚バランス制御を行う.
  \begin{align}
    d &= F_{lhip}-F_{rhip}\\
    D &\gets D + d\\
    \theta_{T-r} &= C_{pgain}d + C_{igain}D
  \end{align}
  ここで, $C_{\{pgain, igain\}}$はPI制御のゲインである.
  これにより, 身体の横ズレを防ぐことが可能である.
  本研究では, $C_{pgain}=5.0$, $C_{igain}$については, 前進後退では$C_{igain}=0.3$, 回旋では$C_{igain}=0.03$とした.
}%

\begin{figure}[t]
  \centering
  \includegraphics[width=0.95\columnwidth]{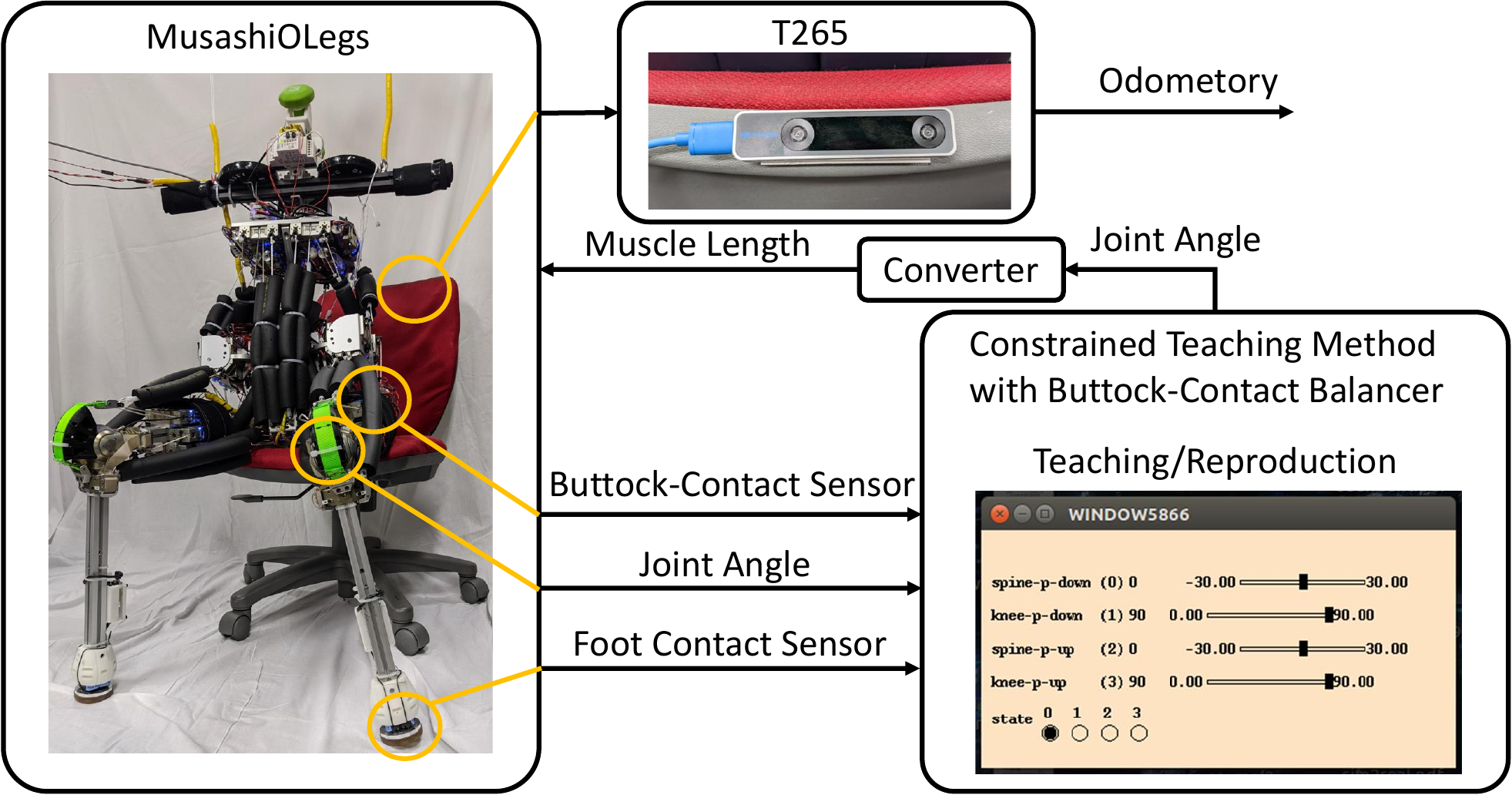}
  \caption{The whole system of the constrained teaching method and buttock-contact balancer.}
  \label{figure:whole-system}
\end{figure}

\subsection{Whole System For Seated Walk} \label{subsec:whole-system}
\switchlanguage%
{%
  The whole system of this study is shown in \figref{figure:whole-system}.
  MusashiOLegs sits on a chair with casters and a swivel seat, and target motion is taught and reproduced by CTM.
  During the movement, the buttock-contact balance control is run simultaneously.
  In order to quantitatively observe the movement, Intel Realsense T265 is attached to the back of the chair and the motion trajectory is measured by visual SLAM.
  The joint angles are converted into muscle lengths by \cite{kawaharazuka2019longtime} and sent to the actual robot.
  The period of the CTM and balance control is set to 5 Hz.

  The control and transition condition functions to be registered in CTM and their transitions for motions of Move-Forward, Move-Backward, Rotate-Left, and Rotate-Right of seated walk are shown below.

  Move-Forward is as follows.
  \begin{enumerate}
    \item $f_1(u_1): \theta_{T-p} = u_1$, $g_1(s_1=F_{foot}, C^{thre}_{1}): s_1 \leq C^{thre}_{1}$
    \item $f_2(u_2): \theta_{K-p} = u_2$, $g_2(s_2=\theta_{lK-p}, C^{thre}_{2}): s_2 \leq C^{thre}_{2}$
    \item $f_3(u_3): \theta_{T-p} = u_3$, $g_3(s_3=F_{foot}, C^{thre}_{3}): s_3 \geq C^{thre}_{3}$
    \item $f_4(u_4): \theta_{K-p} = u_4$, $g_4(s_4=\theta_{lK-p}, C^{thre}_{4}): s_4 \geq C^{thre}_{4}$
  \end{enumerate}
  Note that $\theta_{K-p}=u$ means that $\theta_{lK-p}=u$ and $\theta_{rK-p}=u$ are performed simultaneously.
  The motion is to bend at the waist, raise the legs, round at the waist to put the weight on the legs, and bend the knees to move forward.
  We set $\Delta{u} = \{u_1, u_2, \cdots, u_{N^{state}}\} = \{-2, -3, 2, 1\}$ [deg].

  Move-Backward is as follows.
  \begin{enumerate}
    \item $f_1(u_1): \theta_{T-p} = u_1$, $g_1(s_1=F_{foot}, C^{thre}_{1}): s_1 \leq C^{thre}_{1}$
    \item $f_2(u_2): \theta_{K-p} = u_2$, $g_2(s_2=\theta_{K-p}, C^{thre}_{2}): s_2 \geq C^{thre}_{2}$
    \item $f_3(u_3): \theta_{T-p} = u_3$, $g_3(s_3=F_{foot}, C^{thre}_{3}): s_3 \geq C^{thre}_{3}$
    \item $f_4(u_4): \theta_{K-p} = u_4$, $g_4(s_4=\theta_{K-p}, C^{thre}_{4}): s_4 \leq C^{thre}_{4}$
  \end{enumerate}
  The motion is to bend at the waist, lower the legs, round at the waist to put the weight on the legs, and raise the knees to move backward.
  We set $\Delta{u} = \{-2, 3, 2, -1\}$ [deg].

  Rotate-Left is as follows.
  \begin{enumerate}
    \item $f_1(u_1): \theta_{lH-p} = u_1$, $g_1(s_1=F_{lfoot}, C^{thre}_{1}): s_1 \leq C^{thre}_{1}$
    \item $f_2(u_2): \theta_{H-r} = u_2$, $g_2(s_2=\theta_{lH-r}, C^{thre}_{2}): s_2 \geq C^{thre}_{2}$
    \item $f_3(u_3): \theta_{lH-p} = u_3$, $g_3(s_3=F_{lfoot}, C^{thre}_{3}): s_3 \geq C^{thre}_{3}$
    \item $f_4(u_4): \theta_{rH-p} = u_4$, $g_4(s_4=F_{rfoot}, C^{thre}_{4}): s_4 \leq C^{thre}_{4}$
    \item $f_5(u_5): \theta_{H-r} = u_5$, $g_5(s_5=\theta_{lH-r}, C^{thre}_{5}): s_5 \leq C^{thre}_{5}$
    \item $f_6(u_6): \theta_{rH-p} = u_6$, $g_6(s_6=F_{rfoot}, C^{thre}_{6}): s_6 \geq C^{thre}_{6}$
  \end{enumerate}
  Note that $\theta_{H-r}=u$ means that $\theta_{lH-r}=u$ and $\theta_{rH-r}=-u$ are performed simultaneously.
  The motion is to raise the left leg, open the crotch for left rotation, lower the left leg, raise the right leg, close the crotch, and lower the right leg.
  We set $\Delta{u} = \{-2, 2, 2, -2, -2, 2\}$ [deg].

  Rotate-Right is as follows.
  \begin{enumerate}
    \item $f_1(u_1): \theta_{rH-p} = u_1$, $g_1(s_1=F_{rfoot}, C^{thre}_{1}): s_1 \leq C^{thre}_{1}$
    \item $f_2(u_2): \theta_{H-r} = u_2$, $g_2(s_2=\theta_{lH-r}, C^{thre}_{2}): s_2 \geq C^{thre}_{2}$
    \item $f_3(u_3): \theta_{rH-p} = u_3$, $g_3(s_3=F_{rfoot}, C^{thre}_{3}): s_3 \geq C^{thre}_{3}$
    \item $f_4(u_4): \theta_{lH-p} = u_4$, $g_4(s_4=F_{lfoot}, C^{thre}_{4}): s_4 \leq C^{thre}_{4}$
    \item $f_5(u_5): \theta_{H-r} = u_5$, $g_5(s_5=\theta_{lH-r}, C^{thre}_{5}): s_5 \leq C^{thre}_{5}$
    \item $f_6(u_6): \theta_{lH-p} = u_6$, $g_6(s_6=F_{lfoot}, C^{thre}_{6}): s_6 \geq C^{thre}_{6}$
  \end{enumerate}
  The motion is to raise the right leg, open the crotch for right rotation, lower the right leg, raise the left leg, close the crotch, and lower the left leg.
  We set $\Delta{u} = \{-2, 2, 2, -2, -2, 2\}$ [deg].
}%
{%
  本研究の全体システムを\figref{figure:whole-system}に示す.
  キャスター付きで座面が回転する椅子にMusashiOLegsが座り, constrained teaching methodにより動作を教示・再生する.
  動作時, お尻接触バランス制御を同時に走らせておく.
  このとき, 移動の具合を定量的に見るために, Intel Realsense T265を椅子の後ろに取り付け, visual slamによって移動した経路を測定する.
  関節角度は\cite{kawaharazuka2019longtime}により筋長に変換されて実機に送られる.
  なお, CTMやバランス制御の周期は5 Hzとした.

  seated walkのMove-Forward, Move-Backward, Rotate-Left, Rotate-Rightにおける, Constrained Teaching Methodに登録する制御・条件関数とその遷移を示す.
  Move-Forwardは以下である.
  \begin{enumerate}
    \item $f_0(u_0): \theta_{T-p} = u_0$, $g_0(s_0=F_{foot}, C^{thre}_{0}): s_0 \leq C^{thre}_{0}$
    \item $f_1(u_1): \theta_{K-p} = u_1$, $g_1(s_1=\theta_{lK-p}, C^{thre}_{1}): s_1 \leq C^{thre}_{1}$
    \item $f_2(u_2): \theta_{T-p} = u_2$, $g_2(s_2=F_{foot}, C^{thre}_{2}): s_2 \geq C^{thre}_{2}$
    \item $f_3(u_3): \theta_{K-p} = u_3$, $g_3(s_3=\theta_{lK-p}, C^{thre}_{3}): s_3 \geq C^{thre}_{3}$
  \end{enumerate}
  なお, $\theta_{K-p}=u$は, $\theta_{lK-p}=u$と$\theta_{rK-p}=u$を同時に行うことを意味する.
  腰を反らして足を上げ, 腰を丸めて足に体重をかけ, 膝を曲げて前に進む動作とした.
  再生時は$\Delta{u} = \{u_1, u_2, \cdots, u_{N^{state}}\} = \{-2, -3, 2, 1\}$ [deg]とした.

  Move-Backwardは以下である.
  \begin{enumerate}
    \item $f_0(u_0): \theta_{T-p} = u_0$, $g_0(s_0=F_{foot}, C^{thre}_{0}): s_0 \leq C^{thre}_{0}$
    \item $f_1(u_1): \theta_{K-p} = u_1$, $g_1(s_1=\theta_{K-p}, C^{thre}_{1}): s_1 \geq C^{thre}_{1}$
    \item $f_2(u_2): \theta_{T-p} = u_2$, $g_2(s_2=F_{foot}, C^{thre}_{2}): s_2 \geq C^{thre}_{2}$
    \item $f_3(u_3): \theta_{K-p} = u_3$, $g_3(s_3=\theta_{K-p}, C^{thre}_{3}): s_3 \leq C^{thre}_{3}$
  \end{enumerate}
  腰を反らして足を下げ, 腰を丸めて足に体重をかけ, 膝を上げて後ろに進む動作とした.
  再生時は$\Delta{u} = \{-2, 3, 2, -1\}$ [deg]とした.

  Rotate-Leftは以下である.
  \begin{enumerate}
    \item $f_0(u_0): \theta_{lH-p} = u_0$, $g_0(s_0=F_{lfoot}, C^{thre}_{0}): s_0 \leq C^{thre}_{0}$
    \item $f_1(u_1): \theta_{H-r} = u_1$, $g_1(s_1=\theta_{lH-r}, C^{thre}_{1}): s_1 \geq C^{thre}_{1}$
    \item $f_2(u_2): \theta_{lH-p} = u_2$, $g_2(s_2=F_{lfoot}, C^{thre}_{2}): s_2 \geq C^{thre}_{2}$
    \item $f_3(u_3): \theta_{rH-p} = u_3$, $g_3(s_3=F_{rfoot}, C^{thre}_{3}): s_3 \leq C^{thre}_{3}$
    \item $f_4(u_4): \theta_{H-r} = u_4$, $g_4(s_4=\theta_{lH-r}, C^{thre}_{4}): s_4 \leq C^{thre}_{4}$
    \item $f_5(u_5): \theta_{rH-p} = u_5$, $g_5(s_5=F_{rfoot}, C^{thre}_{5}): s_5 \geq C^{thre}_{5}$
  \end{enumerate}
  なお, $\theta_{H-r}=u$は, $\theta_{lH-r}=u$と$\theta_{rH-r}=-u$を同時に行うことを意味する.
  左足を上げて股を開いて左回旋し, 左足を下げて右足を上げて股を閉じ, 右足を下げるような動作とした.
  再生時は$\Delta{u} = \{-2, 2, 2, -2, -2, 2\}$ [deg]とした.

  Rotate-Rightは以下である.
  \begin{enumerate}
    \item $f_0(u_0): \theta_{rH-p} = u_0$, $g_0(s_0=F_{rfoot}, C^{thre}_{0}): s_0 \leq C^{thre}_{0}$
    \item $f_1(u_1): \theta_{H-r} = u_1$, $g_1(s_1=\theta_{lH-r}, C^{thre}_{1}): s_1 \geq C^{thre}_{1}$
    \item $f_2(u_2): \theta_{rH-p} = u_2$, $g_2(s_2=F_{rfoot}, C^{thre}_{2}): s_2 \geq C^{thre}_{2}$
    \item $f_3(u_3): \theta_{lH-p} = u_3$, $g_3(s_3=F_{lfoot}, C^{thre}_{3}): s_3 \leq C^{thre}_{3}$
    \item $f_4(u_4): \theta_{H-r} = u_4$, $g_4(s_4=\theta_{lH-r}, C^{thre}_{4}): s_4 \leq C^{thre}_{4}$
    \item $f_5(u_5): \theta_{lH-p} = u_5$, $g_5(s_5=F_{lfoot}, C^{thre}_{5}): s_5 \geq C^{thre}_{5}$
  \end{enumerate}
  右足を上げて股を開いて右回旋し, 右足を下げて左足を上げて股を閉じ, 左足を下げるような動作とした.
  再生時は$\Delta{u} = \{-2, 2, 2, -2, -2, 2\}$ [deg]とした.
}%

\begin{figure*}[t]
  \centering
  \includegraphics[width=1.95\columnwidth]{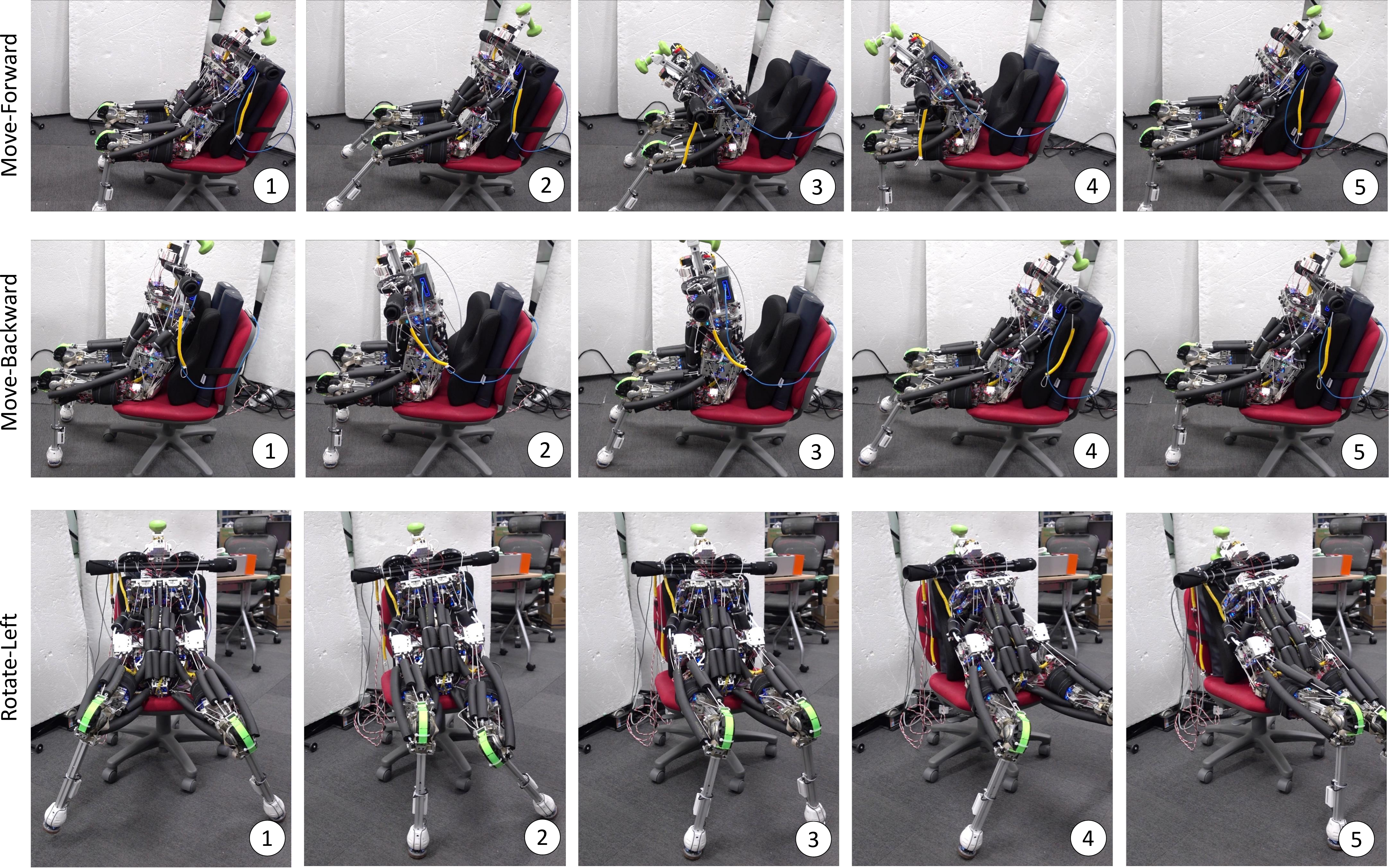}
  \caption{Experiments of Move-Forward, Move-Backward, and Rotate-Left of seated walk by MusashiOLegs.}
  \label{figure:basic-exp}
\end{figure*}

\section{Experiments} \label{sec:experiments}
\subsection{Forward, Backward, and Rotational Movements} \label{subsec:move-exp}
\switchlanguage%
{%
  We show the experimental results of seated walk by MusashiOLegs.
  We performed Move-Forward, Move-Backward, and Rotate-Left (only Rotate-Left was performed because the left and right rotations were symmetrical) in the order of teaching and reproduction.
  Each motion is shown in \figref{figure:basic-exp}.
  By learning only the threshold of the transition condition, each motion was successfully performed.
  The measurement by visual SLAM showed that the robot moved forward by 0.20 m, backward by 0.15 m, and rotated by 23 deg in one state transition loop.

  The thresholds of the transition conditions obtained were $C^{thre}_0=0$ [N], $C^{thre}_1=51.3$ [deg], $C^{thre}_2=75.8$ [N], and $C^{thre}_3=90$ [deg] for Move-Forward.
  The motion of bending backward at the waist until the legs are completely apart from the ground, extending the knees, bending forward at the waist until a force of 76 N is applied to both legs, and bending the knees to move forward was reproduced.
  In Move-Backward, $C^{thre}_0=0$ [N], $C^{thre}_1=90$ [deg], $C^{thre}_2=6.3$ [N], and $C^{thre}_3=46.8$ [deg].
  It was possible to move backward by applying a force of only 6.3 N to both legs, which is much smaller than that for Move-Forward.
  In Move-Forward, when $C^{thre}_2=38.0$ [N], the foot slipped and it was not possible to move forward.
  In Rotate-Left, $C^{thre}_0=0$ [N], $C^{thre}_1=30.4$ [deg], $C^{thre}_2=1.92$ [N], $C^{thre}_3=0.0$ [deg], $C^{thre}_4=4.0$ [deg], and $C^{thre}_5=5.24$ [N].

  The motion durations of teaching and reproduction were 47 [sec] $\rightarrow$ 17 [sec] for Move-Forward, 42 [sec] $\rightarrow$ 16 [sec] for Move-Backward, and 63 [sec] $\rightarrow$ 17 [sec] for Rotate-Left.
  This shows the advantage of learning only the threshold of the transition condition and being able to change the execution time arbitrarily.
}%
{%
  MusashiOLegsによるseated walkの実験結果を示す.
  前進・後退・回旋(左回旋と右回旋は対称のため, 左回旋のみ行う)について, 教示・再生の順に動作を行った.
  \figref{figure:basic-exp}にそれぞれの動作を示す.
  遷移条件の閾値のみを学習することで, それぞれの動作が成功した.
  Visual Slamによる計測では, 一回の状態遷移ループで, 前進は0.20 m, 後退は0.15 m, 回旋は23 deg, 進んでいることがわかった.

  得られた遷移条件の閾値は, 前進では$C^{thre}_0=0$ [N], $C^{thre}_1=51.3$ [deg], $C^{thre}_2=75.8$ [N], $C^{thre}_3=90$ [deg]であった.
  足が完全に離れるまで腰を後ろに倒してから膝を伸ばし, 両足に76 N程度まで力がかかってから膝を曲げて前進するような動作が再生された.
  後退では$C^{thre}_0=0$ [N], $C^{thre}_1=90$ [deg], $C^{thre}_2=6.3$ [N], $C^{thre}_3=46.8$ [deg]であった.
  前進に比べて小さな6.3 Nのみの力をかけるだけで後退が可能であった.
  なお, 前進において$C^{thre}_2=38.0$ [N]のときは, 足が滑って前進することはできなかった.
  回旋では, $C^{thre}_0=0$ [N], $C^{thre}_1=30.4$ [deg], $C^{thre}_2=1.92$ [N], $C^{thre}_3=0.0$ [deg], $C^{thre}_4=4.0$ [deg], $C^{thre}_5=5.24$ [N]であった.

  教示時$\rightarrow$再生時の動作時間は, 前進で47 [sec] $\rightarrow$ 17 [sec], 後退で42 [sec] $\rightarrow$ 16 [sec], 回旋で 63 [sec] $\rightarrow$ 17 [sec]であった.
  状態遷移の閾値のみ学習し, 実行時間を任意に変更できることの利点が示されている.
}%

\begin{figure}[t]
  \centering
  \includegraphics[width=0.95\columnwidth]{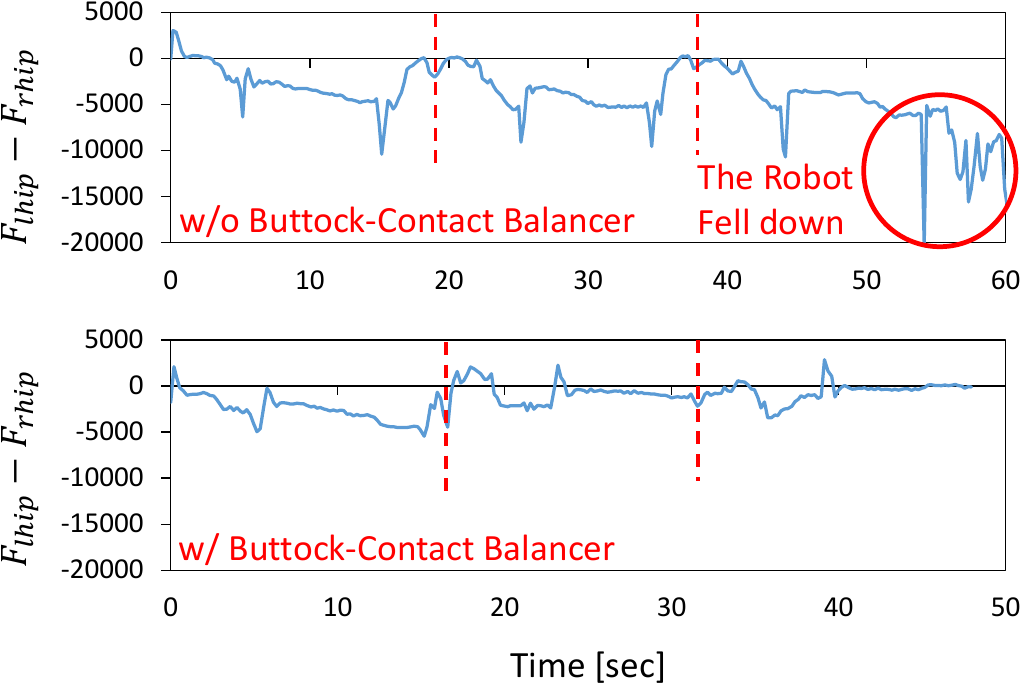}
  \caption{Transition of $F_{lhip}-F_{rhip}$ when moving forward by seated walk with and without buttock-contact balancer.}
  \label{figure:move-balancer-exp}
\end{figure}

\begin{figure}[t]
  \centering
  \includegraphics[width=0.95\columnwidth]{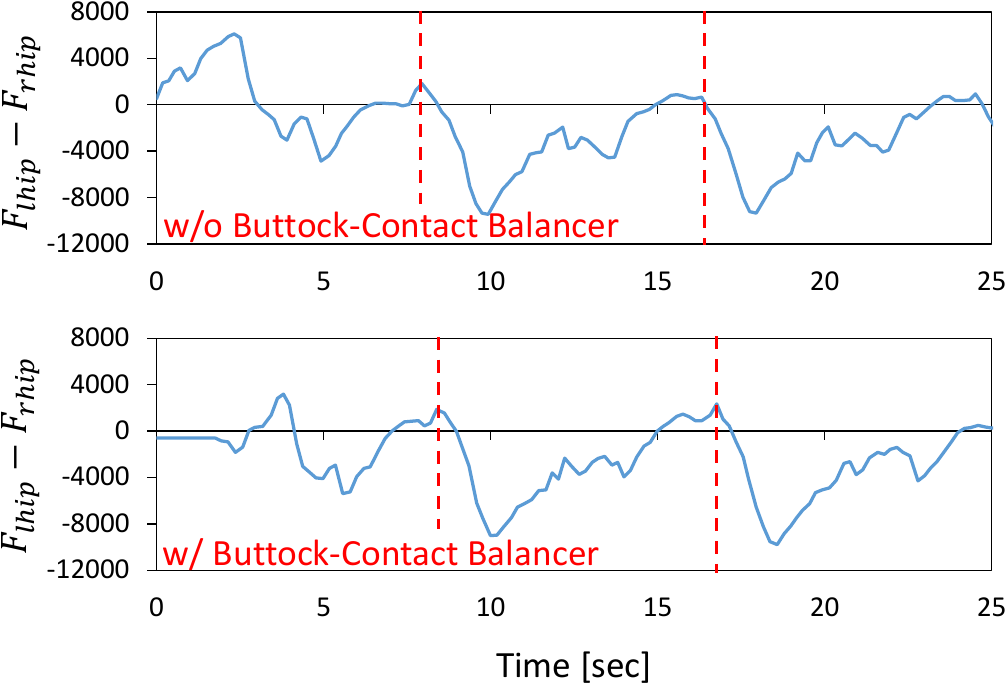}
  \caption{Transition of $F_{lhip}-F_{rhip}$ when rotating left by seated walk with and without buttock-contact balancer.}
  \label{figure:rotate-balancer-exp}
\end{figure}

\begin{figure*}[t]
  \centering
  \includegraphics[width=1.95\columnwidth]{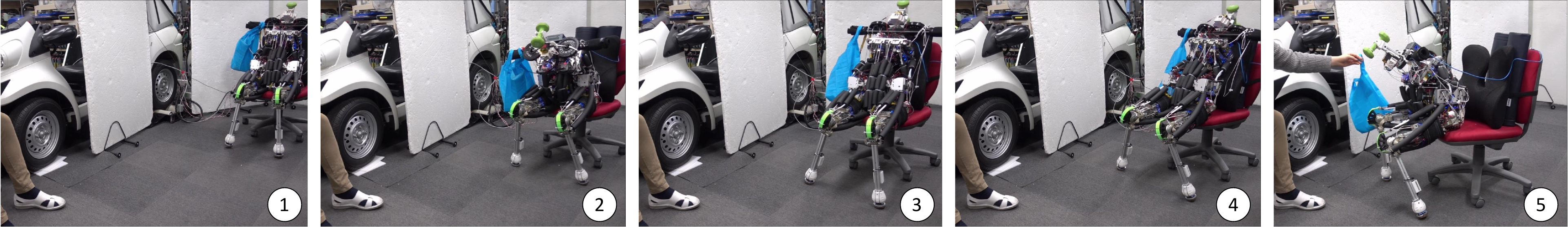}
  \caption{An experiment of carrying a bag by seated walk.}
  \label{figure:complete-exp}
\end{figure*}

\begin{figure}[t]
  \centering
  \includegraphics[width=0.95\columnwidth]{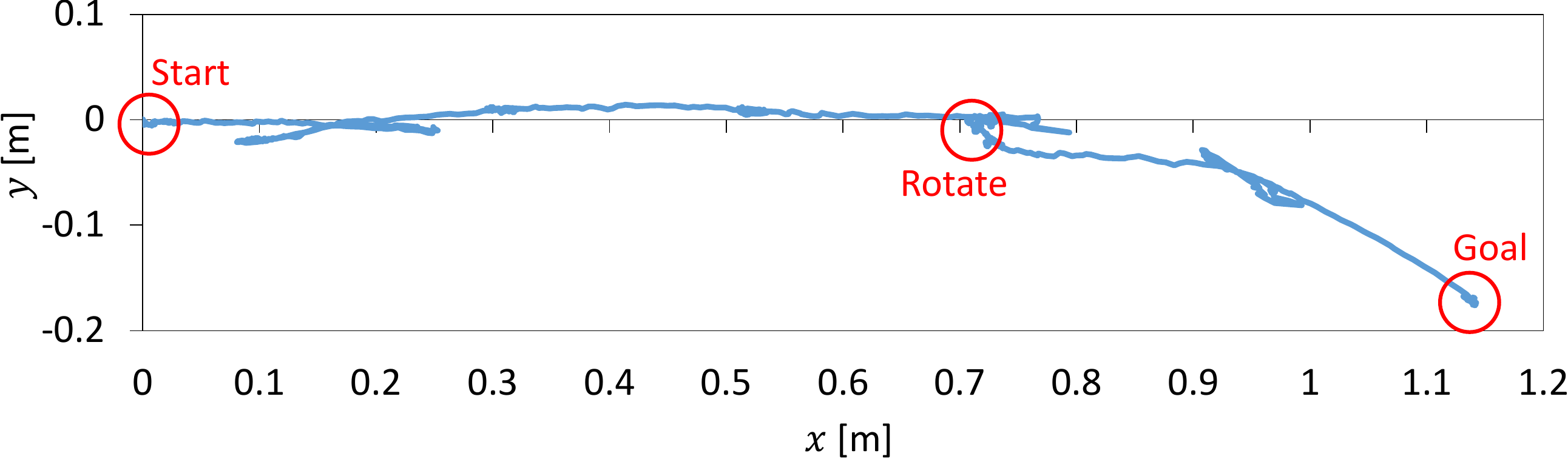}
  \caption{The trajectory of the robot when carrying a bag by seated walk.}
  \label{figure:complete-traj-exp}
\end{figure}

\subsection{Buttock-Contact Balancer} \label{subsec:balancer-exp}
\switchlanguage%
{%
  We show the difference in translational and rotational motions with and without the buttock-contact balancer.
  \figref{figure:move-balancer-exp} shows $F_{lhip}-F_{rhip}$ in Move-Forward, and \figref{figure:rotate-balancer-exp} shows $F_{lhip}-F_{rhip}$ in Rotate-Left.
  Each experiment represents the results of the same motions performed three times consecutively.
  Note that the time taken for each motion is different because of only looking at the threshold of the sensor.
  In Move-Forward, the pressure difference between the left and right buttocks was kept constant with buttock-contact balancer, while the pressure difference increased gradually without it, and the robot fell down in the third movement.
  For Rotate-Left, there was no significant difference in the pressure transition.
}%
{%
  Buttock-Contact Balancerを入れた時と入れない時で, 並進・回転動作に置いて, どのような違いが出るかを示す.
  \figref{figure:move-balancer-exp}には前進における$F_{lhip}-F_{rhip}$を, \figref{figure:rotate-balancer-exp}には左回旋における$F_{lhip}-F_{rhip}$を示す.
  なお, それぞれの実験は同じ動作を3回連続で行った結果を現している.
  センサの閾値のみを見ているため毎回の動作においてかかる時間は異なることに注意されたい.
  前進については, balancerを入れた場合は左右のお尻の圧力差が一定に保たれているのに対して, balancerを入れない場合は徐々に圧力差が増え, 3回目の動作で転倒してしまった.
  左回旋については, 圧力の遷移について大きな差は見られなかった.
}%

\subsection{Carrying a Bag by MusashiOLegs} \label{subsec:carrying-exp}
\switchlanguage%
{%
  Finally, we conducted an experiment in which a bag was delivered by combining translational and rotational movments.
  The threshold values of the transition conditions were obtained by human teaching in advance.
  The experiment is shown in \figref{figure:complete-exp}.
  The robot successfully delivered the blue bag by moving forward four times, rotating to the right once, and moving forward twice.
  The trajectory of the robot is shown in \figref{figure:complete-traj-exp}, and the robot was able to move 1.14 m in the x direction and -0.18 m in the y direction in 145 seconds.
  By combining the motions generated by our method, more complex motions are possible.
}%
{%
  最後に, 並進・回転を合わせて, バックを届ける実験を行った.
  予め教示により遷移条件の閾値を獲得しておき, それを操作者が組み合わせることでかばんを運んだ.
  \figref{figure:complete-exp}にその様子を示す.
  前進を4回, 右回旋を1回, 前進を2回順に行うことで, 青いバックを届けることに成功した.
  その際のロボットの軌跡は\figref{figure:complete-traj-exp}であり, 145秒間でx方向に1.14 m, y方向に-0.18 m進むことができた.
  本手法により生成された動作を組み合わせることとで, より複雑な動作が可能となる.
}%

\section{Discussion} \label{sec:discussion}
\switchlanguage%
{%
  From the experiments in this study, we found that the constrained teaching method can realize the forward, backward, and rotational movements of seated walk, and that the reproduced motion can be executed much faster than the taught motion depending on the control speed setting.
  In addition, it was found that the buttock-contact balance control is significantly important for the translational direction of forward and backward movements, but did not make a significant difference for the rotational movement.
  It was also found that the robot can move in arbitrary directions by combining the taught motions of seated walk.
  Although there are no good examples of three-dimensional walking in flexible musculoskeletal humanoids, the seated walk enables the robot to greatly expand the range of movement.

  On the other hand, there are many limitations in this research, especially in the constrained teaching method.
  First of all, the control and transition condition functions must be written down by humans, which takes more time to prepare than the ordinary teaching.
  We would like to develop a method that estimates and extracts explicit state transitions from human teaching, and obtains sensor values to be used as control commands and sensor thresholds in each state.
  Second, we assume that only quasi-static states can be handled and that the control command is only one-dimensional.
  Of course, dynamic motions are not the focus of this research, but it would be interesting to mix some of ordinary teaching methods with this research, because this research cannot be applied to the motions where the robot needs to move dynamically even for a moment or move multiple joints in different ways.
  Finally, in this study, odometry information is only used for the evaluation of experiments.
  In the future, it will be more practical if we can find out how to reach a target position on the map by combining the taught actions, and if we can parameterize the taught actions themselves and adjust the magnitude of the movements.
}%
{%
  本研究の実験から, 制約付き教示手法によって, seated walkの前進後退回旋が実現できること, 教示時に比べて再生時は制御速度の設定次第で大幅に速く動作を実行できることがわかった.
  また, お尻接触バランス制御は前進後退の並進方向については顕著に重要であるが, 回旋運動については大きな差がないことがわかった.
  教示したいくつかの動作をつなげて任意方向への移動が可能であることもわかった.
  柔軟な筋骨格ヒューマノイドにおける三次元歩行の実現例はこれまでにないが, seated walkの実現により, 移動範囲を大きく拡大することが可能となった.

  一方で, 本研究, 特にCTMには多くのlimitationが存在する.
  まず, 制御入力・遷移条件を人間が書き下す必要があるため, 通常の教示に比べると準備に時間がかかる.
  教示の際の動きから, 明示的な状態遷移を推定して取り出し, それぞれの状態における制御入力・閾値として用いるべきセンサ値を獲得するような方法へと発展させていきたい.
  次に, 静止状態しか扱えないこと, 制御入力が一次元のみであるという仮定である.
  もちろん動的動作は本研究のフォーカスではないが, 一瞬でも動的に動かしたり複数関節をそれぞれ異なる形で動かす必要がある遷移状態が入ると本研究は適用できなくなってしまうため, 一部だけ通常の教示手法を混ぜて適用できるようにするといったことも面白い.
  最後に, 本研究では, オドメトリ情報を実験の評価にしか用いていない.
  今後, マップ上で与えられた位置に, 教示した動作をどう組み合わせることで到達することができるのか, また, 教示動作自体をパラメータ化し, 動きの大きさも調節できるようになるとさらに実用性が高まると考える.
}%

\section{Conclusion} \label{sec:conclusion}
\switchlanguage%
{%
  In this study, we described a method to achieve seated walk, which has not been realized so far, by using musculoskeletal humanoids that are more human-like and difficult to modelize.
  By implementing a buttock-contact sensor on the planar interskeletal structure that mimics the gluteus maximus muscle, the robot can measure the pressure between the buttocks and the chair and execute balance control.
  In addition, we have developed a constrained teaching method and have succeeded in realizing forward, backward, and rotational movements by learning the threshold of the transition condition from human teaching.
  By narrowing down the control command to one dimension and providing the control and transition condition functions in advance, only the threshold value is learned and the execution speed of reproduction can be arbitrarily changed from human teaching.
  For the first time, we succeeded in the seated walk motion and showed that it is possible for musculoskeletal humanoids to carry objects by combining translation and rotation.
  In the future, we would like to extend this research further and construct a whole system that integrates manipulation by upper limbs and navigation with this study.
}%
{%
  本研究では, より人間に近くモデル化の難しい筋骨格ヒューマノイドを用いて, これまで実現されてこなかったseated walkを達成する方法について述べた.
  筋骨格構造における大殿筋を模倣した面状骨格間構造にお尻接触センサを実装することで, 椅子との間の圧力を測定し, バランス制御を実行することができる.
  また, 準制的な制約付き教示手法を開発し, 人間の教示から動作遷移条件の閾値を学習することで, 前後移動・左右回転動作を実現することに成功した.
  制御入力を一次元に絞って予め与える, かつ遷移条件の式を予め与えることで, 閾値のみを学習し, 教示時と再生時の実行速度も任意に変更することができる.
  そして, 初めてseated walk動作に成功し, 並進と回転を合わせて移動することが可能なことを示した.
  今後本研究をさらに拡張し, 上肢によるマニピュレーションとの統合・navigationも合わせた全体システムを構築していきたい.
}%

{
  %\footnotesize
  %\small
  %\bibliographystyle{junsrt}
  \bibliographystyle{IEEEtran}
  \bibliography{main}
}

\end{document}